\RequirePackage{fix-cm}
\documentclass[smallextended,twocolumn]{svjour3}          
\smartqed  
\usepackage{graphicx}
\usepackage{mathptmx}
\usepackage{amsmath}
\usepackage{amssymb}
\usepackage{url}
\usepackage[ruled,vlined]{algorithm2e}
\usepackage{multirow}
\usepackage[round]{natbib}
\usepackage{hyperref}
\begin{document}

\title{Robot Perception of Static and Dynamic Objects\\ with an Autonomous Floor Scrubber\thanks{This work has received funding from the European Union's Horizon 2020 research and innovation programme under grant agreement No 645376 (FLOBOT).}
}


\author{Zhi Yan$^{*}$ \and Simon Schreiberhuber$^{*}$ \and Georg Halmetschlager\\ Tom Duckett \and Markus Vincze \and Nicola Bellotto}

\authorrunning{Zhi Yan, Simon Schreiberhuber et al.} 

\institute{
  $^{*}$These authors contributed equally to this work.\\~\\
  Zhi Yan \at
  University of Technology of Belfort-Montb\'eliard (UTBM), France.\\
  \email{zhi.yan@utbm.fr}
  \and
  Simon Schreiberhuber, Georg Halmetschlager, Markus Vincze \at
  Technical University Wien (TU Wien), Austria.\\
  \email{\{schreiberhuber, halmetschlager, vincze\}@acin.tuwien.ac.at}
  \and
  Tom Duckett, Nicola Bellotto \at
  Lincoln Centre for Autonomous Systems Research (L-CAS), University of Lincoln, UK.\\
  \email{\{tduckett, nbellotto\}@lincoln.ac.uk}
}

\date{Received: date / Accepted: date}

\maketitle

\begin{abstract}
  This paper presents the perception system of a new professional cleaning robot for large public places.
  The proposed system is based on multiple sensors including 3D and 2D lidar, two RGB-D cameras and a stereo camera.
  The two lidars together with an RGB-D camera are used for dynamic object (human) detection and tracking, while the second RGB-D and stereo camera are used for detection of static objects (dirt and ground objects).
  A learning and reasoning module for spatial-temporal representation of the environment based on the perception pipeline is also introduced.
  Furthermore, a new dataset collected with the robot in several public places, including a supermarket, a warehouse and an airport, is released.
  Baseline results on this dataset for further research and comparison are provided.
  The proposed system has been fully implemented into the Robot Operating System (ROS) with high modularity, also publicly available to the community.
\end{abstract}

\section{Introduction}
\label{intro}

Many industrial, commercial and public buildings, such as supermarkets, airports, trade fairs and hospitals, have huge floor surfaces that need to be cleaned on a daily basis.
Cleaning these surfaces is time-consuming and requires substantial human effort involving repetitive actions.
These cleaning activities take place at different times of the day, often with a tight schedule, depending on the area that has to be cleaned and on the available time slots.
The economic viability of the cleaning service provider often relies on low wages and low-skilled personnel.
Furthermore, cleaning tasks have often been related to workers' health issues.
Therefore, floor washing activities are well-suited to robotic automation.

\begin{figure}[t]
  \centering
  \includegraphics[width=\columnwidth]{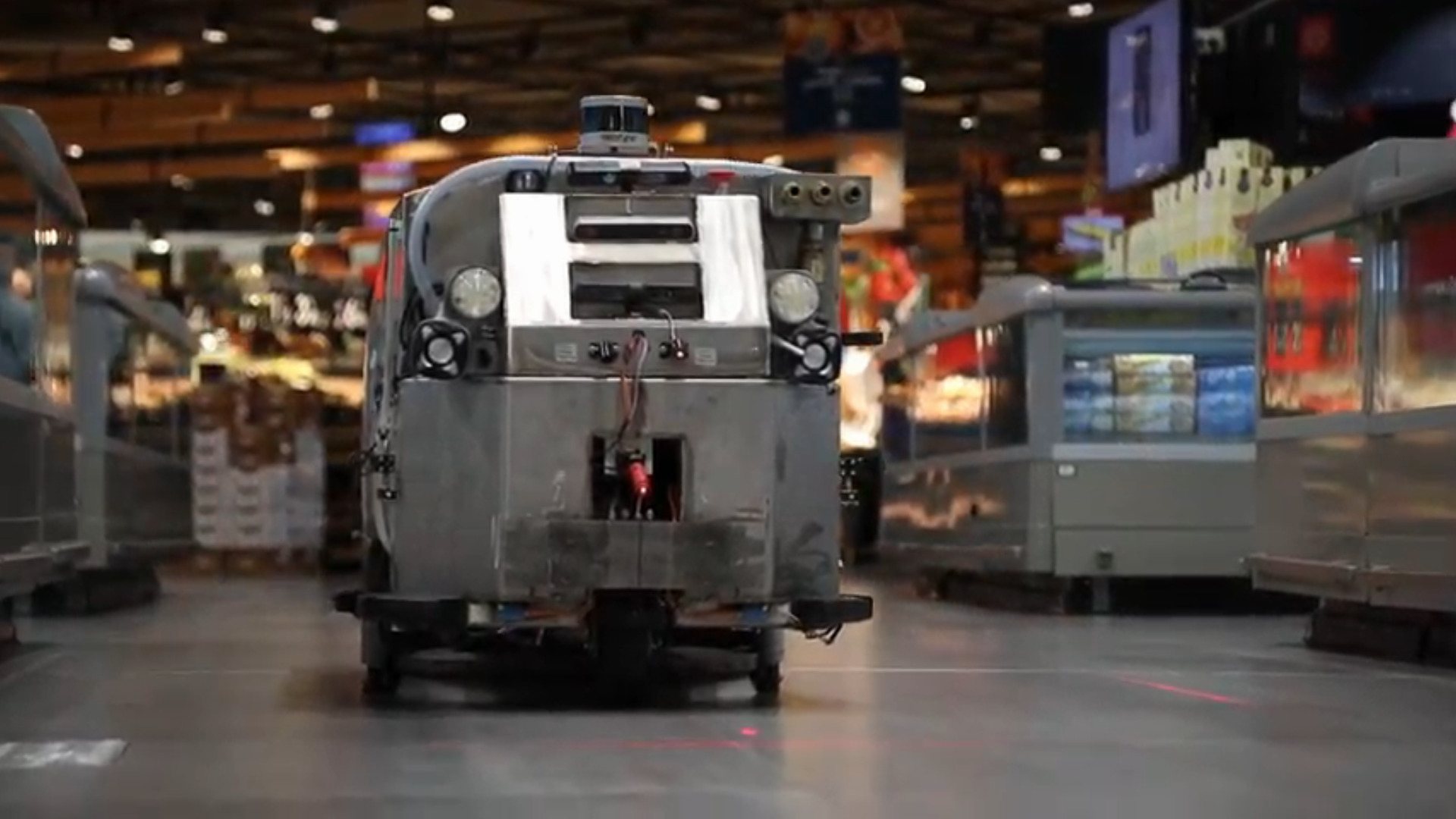}
  \caption{The FLOBOT prototype in action in a supermarket in Italy.}
  \label{fig:flobot}
\end{figure}

However, the development of such a floor washing robot faces many new challenges, including operational autonomy, navigation precision, safety with regards to humans and goods, interaction with the human cleaning personnel, path optimization, easy set-up and reprogramming.
Prior to the EU-funded project FLOBOT (Floor Washing Robot for Professional Users\footnote{\url{http://www.flobot.eu/}}, see Fig.~\ref{fig:flobot}), there was no robot that satisfies the requirements of both professional users and cleaning service providers.

In this paper, we describe the entire perception pipeline of FLOBOT, including software modules for  visual floor inspection and human tracking to enable safe operation.
In addition, the extension of these two modules for learning of and reasoning about the environment surrounding the robot is also presented.
In particular, we use a 3D lidar, an RGB-D camera and a 2D lidar for human detection and tracking, and a stereo camera and a second RGB-D camera for floor dirt and object detection.
The proposed system covers both dynamic (mostly human) and static objects, providing the required perception technologies for robotic cleaning in public spaces.

The contributions of this paper are four-fold.
First, we present a large-scale (long-range and wide-angle) human detection and tracking system using three heterogeneous sensors.
A high-level fusion method uses data association algorithms to combine the detections from each sensor.
The proposed system also includes a new RGB-D camera-based leg detector.
Second, we introduce a new online method to detect ground dirt in front of the robot without the need for pre-training on dirt and floor samples.
Third, we cumulatively gather the information about the dynamic and static objects during the robot's work process, building and refining a spatial-temporal model of the environment, and develop high-level semantics which can help to improve future cleaning schedules.
Fourth, we introduce a new dataset accessible for public download\footnote{\url{http://lcas.github.io/FLOBOT/}}, entirely based on ROS (Robot Operating System)~\citep{ros}, which was collected with the real robot prototype in real environments including an airport, warehouse and supermarket.
These data are difficult to obtain, and similar datasets were previously unavailable to the research community.

The remainder of this paper is organized as follows.
Sect.~\ref{sec:related_work} gives an overview of the related literature.
Then, we introduce the FLOBOT perception system in Sect.~\ref{sec:system}, including both hardware and software aspects.
Sect.~\ref{sec:dynamic_perception}, \ref{sec:static_perception} and \ref{sec:reasoning_learning} detail the human detection and tracking, dirt and object detection, as well as the environment reasoning and learning modules, respectively.
Sect.~\ref{sec:evaluation} presents our dataset and the corresponding evaluation results for our system.
Finally, conclusions and future research directions are discussed in Sect.~\ref{sec:conclusions}.

\section{Related work}
\label{sec:related_work}

\subsection{Human detection and tracking}

Human detection and tracking are essential for service robots, as a robot often shares its workspace and interacts closely with humans.
As FLOBOT uses a 3D lidar, an RGB-D camera and a 2D lidar for human detection and tracking, we first review some related work using single sensors, followed by a discussion of methods fusing data from multiple sensors.

3D lidar has been adopted by a growing number of researchers and industries, thanks to its ability to provide accurate geometrical (point cloud) information about its environment over a long range and wide angle.
However, due to the low feature density compared to cameras, false positives are more likely.
The situation is even worse when the person is far away from the sensor as the point cloud becomes increasingly sparse with distance.

Existing work on 3D-lidar-based human detection can be roughly divided into two categories, namely segmentation-classification pipelines and end-to-end pipelines.
The former first clusters the point cloud~\citep{yz19auro,yz17iros,zermas17icra,bogoslavskyi16iros} then classifies the cluster based on a given model.
This model can be based on machine learning~\citep{yz19auro,yz17iros,kidono11iv} or object motion~\citep{dewan16icraMotionCurve,shackleton10avssTrackingDriven}.
The end-to-end pipeline is nowadays closely linked to deep learning methods, which allow us to extract pedestrians and other objects directly from the point cloud~\citep{VoxelNet,YOLO3D}.

The RGB-D camera has been widely used for human detection for many years.
Although the visual range is relatively narrow, it can accurately perform detection and tracking tasks due to its ability to combine color and dense depth information~\citep{spinello11iros}.
Later work has shown that performance can be further improved without sacrificing detection accuracy if we only check the upper body of the person from the depth data using template matching~\citep{jafari14icra}.

So far, 2D lidar is still the most widely used tool for robotic mapping and localization.
However, since usually installed close to the ground, it is also particularly suitable for human leg detection~\citep{arras07icra}.
Although false alarms are difficult to avoid, the 2D lidar can still provide a useful contribution to the robustness of the perception system.

Conventionally, each type of sensor performs a specific function and, only in rare cases, shares information with other sensors.
However, relying solely on a single sensor prevents the  implementation of more advanced and safer navigation algorithms in autonomous mobile service robots for human environments.
A practical and effective multi-sensor-based method was proposed by~\citep{Bellotto2009}.
It combines a monocular camera and a 2D lidar, utilizing a fast implementation of the unscented Kalman filter (UKF) to achieve real-time, robust multi-person tracking.
In order to deal with people tracking for mobile robots in very crowded and dynamic environments, \citep{linder16icra} presented a multi-modal system using two RGB-D cameras, a stereo camera and two 2D lidars.
For outdoor scenarios, \citep{spinello09ijrr} introduced an integrated system to detect and track people and cars using a camera and a 2D lidar installed on an autonomous car, while \citep{kobilarov06icra} mainly focused on fast-moving people tracking.

Despite a thorough review of the prior art, we did not find any related work demonstrating sensor fusion with 3D lidar data for people tracking as FLOBOT does.
\citep{held13icra} developed an algorithm to align 3D lidar data with high-resolution camera images but for vehicle tracking only.
In our previous work~\citep{yz19auro,yz17iros} we illustrated an online learning framework for 3D lidar-based human detection, in~\citep{ls18icra} we showed an efficiency trajectory prediction using deep learning, while in~\citep{yz18iros} an online transfer learning framework is described for 3D lidar-based human detection.

\subsection{Dirt detection}
Cleaning robots have proven to be the pioneers of personal service robots and started to populate our homes.
Although many are based on simple behaviours, there is an increasing trend towards sensor-based systems with awareness of their environment.
But while behaviour-based systems are being augmented by SLAM-driven approaches, awareness of dirt and other pollutants is still not part of any current systems.

The utility of such dirt detection technology lies not only in giving robots the ability to approach cleaning tasks in a proactive fashion.
It would also enable cleaning contractors to quantify their service.
Turbidity sensors were considered and tested for this task since they are already applied in machines like dishwashers, but were not pursued further since we strive for robots that anticipate instead of just react.

There is little work approaching visual dirt detection, and the few methods tackling this task reduce the problem to classification of clean versus polluted areas.
The method proposed by~\citep{rb13icra} assumes different spatial frequencies in the polluted and clean areas of the images.
Effectively the background/floor is therefore limited to only one frequency/color whereas everything outside this spectrum is classified as dirt.
This situation also influences the availability of datasets, of which to our knowledge there is only one~\citep{rb13icra}.

Novelty detection provides a general framework for solving the dirt detection task.
Classical approaches~\citep{pimentel_2014} are often frugal in their data consumption but not as effective as modern CNN based approaches like~\citep{grunwald_2018} which have involved training processes.
We found GMM-based approaches like~\citep{drews_2013} to be quite robust, even when the application is not as well delimited as in~\citep{grunwald_2018}.

\subsection{Object detection}

Detecting objects and evading them is typically part of the navigation module, which often relies solely on lidar data.
A top-mounted 3D lidar often leaves blind spots in the driving direction due to occlusion by the chassis and the limited vertical field of view.
Small objects would therefore only be perceivable at a distance too high for reliable detection. 

In the context of cleaning robots, this could be problematic depending on the utilized cleaning equipment.
For example, with a rotating brush tiny objects could be spun away, which is not necessarily desired.
In the case of the robot only being equipped with a rubber lip (e.g. squeegee), objects could interfere with cleaning operations by jamming between the rubber and floor.
In the case of human driven cleaning machines, this often requires the operator to manually remove the obstacle.

With floor-facing RGB-D or stereo cameras, we have cost-effective options to detect these obstacles and take corresponding actions, especially since fitting a plane model to the floor fits the needs of our scenarios.
Everything protruding above this model with sufficient significance is considered an obstacle.
The most prominent method of fitting such a plane model is RANSAC~\citep{Jia_2018,yang_2010,tarsha_2008}.
Working on disparity images, plane extraction can also be achieved by line fitting in v-disparity space~\citep{dy13iv,zhao_2007}.
In their initial form these algorithms only fit one perfect plane to a given input frame, whereas reality often demands more flexible floor models.

Our work in~\citep{ss17oagm} gives room for some curvature along planes to compensate for inaccuracies in both floor and sensor.
We furthermore adapted the noise model derived by~\citep{ghf19ram} to guide a more sensible thresholding scheme that allows us to detect objects as small as $2cm$ at distances smaller than $1.3m$.

\subsection{Environment reasoning and learning}

To enable a service robot to achieve robust and intelligent behaviour in human environments for extended periods (i.e. long-term autonomy), continuous learning and reasoning about the environment is key~\citep{Kunze2018}.
Pioneering work \citep{fremen} focuses on representing the uncertainty by combination of periodic functions obtained through frequency analysis (i.e. the FreMEn method).
In particular, it models the uncertainties as probabilistic functions of time, allowing integration of long-term observations of the same environment into memory-efficient spatio-temporal models.
To extend the discrete FreMEn framework to both discrete and continuous spatial representations, \citep{hypertime} expanded the spatial model with a set of wrapped time dimensions that represent the periodicities of the observed events.
By using this new representation, \citep{vintr19icra} modeled periodic temporal patterns of people presence, based on peoples' routines and habits, in a human populated environment.
The experimental results showed the capability of long-term predictions of human presence, allowing mobile robots to schedule their services better and to plan their paths.

For professional cleaning robots like FLOBOT serving large public places, both static and dynamic objects in the environment are worth learning.
Different from the previous representations, we use heatmaps to model the presence of humans (dynamics)~\citep{ls18icra}, dirt and static objects~\citep{ag17taros}, in both continuous and discrete spaces.
The heatmap is a graphical representation of data where the individual values contained in a matrix are represented as colours, which can provide an intuitive portrayal of the changing environment.

\section{FLOBOT perception system}
\label{sec:system}

Perception ability is an important feature that distinguishes robots from traditional automata.
Effective perception is an essential component of many modules required for an autonomous robot to operate safely and reliably in our daily life.
FLOBOT is equipped with a variety of advanced sensors to build a heterogeneous and complete sensing system for both internal (e.g. velocity and orientation of the robot) and external (e.g. image and distance of the object) factors.
The requirement for multiple sensors is mainly due to the fact that different sensors have different (physical) properties, and each category has its own strengths and weaknesses~\citep{yz18iros}.
Meanwhile, ROS has become the \emph{de facto} standard platform for development of software in robotics.
Its high modularity and reusability facilitate the cooperative development within the project consortium and the dissemination of results to the community.
Next, we introduce the FLOBOT perception system including both hardware and software aspects.

\subsection{Hardware configuration}

\begin{figure}[t]
  \centering
  \includegraphics[width=\columnwidth]{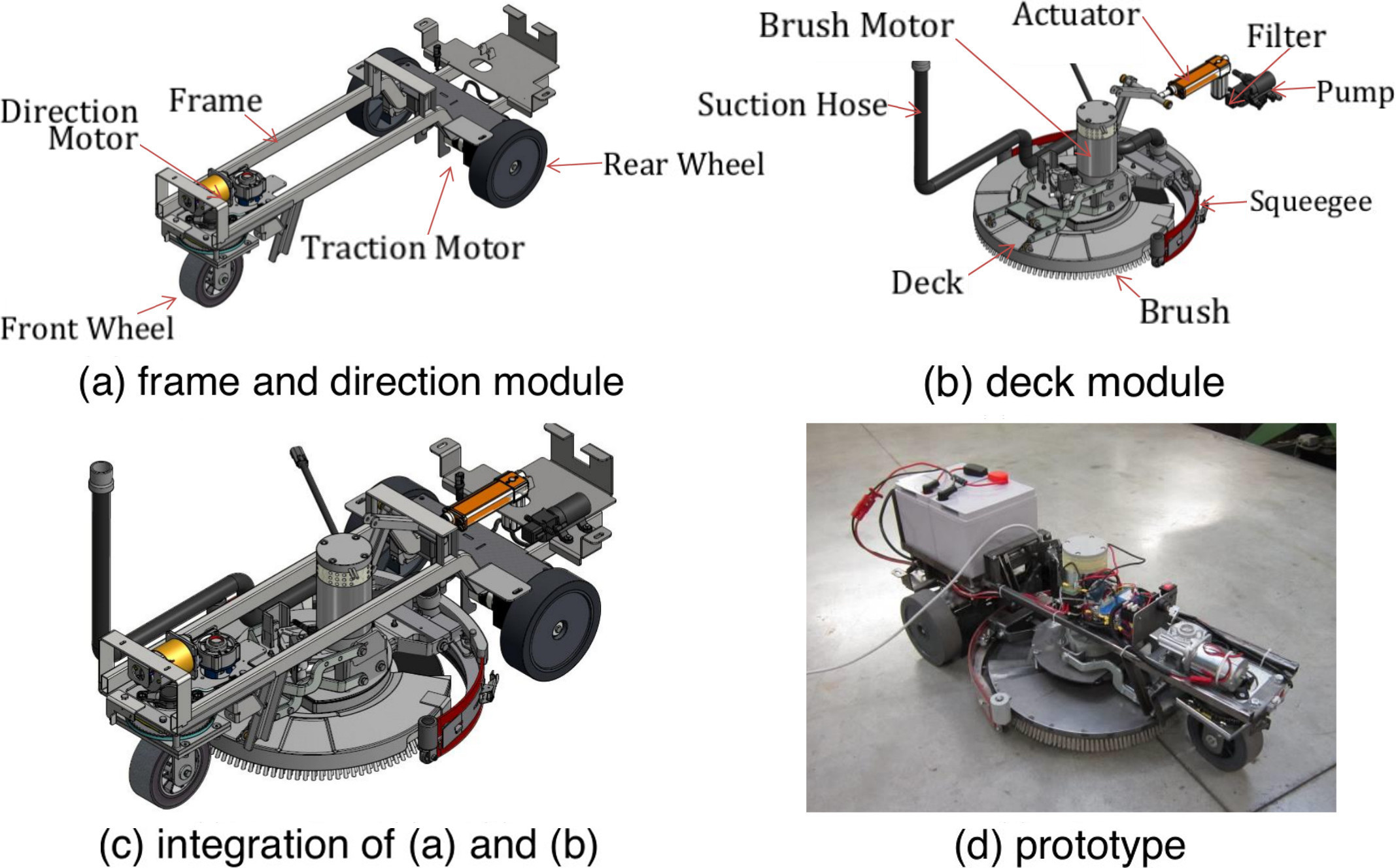}
  \caption{The three-wheeled mobile base and the cleaning unit of FLOBOT prototype.}
  \label{fig:mobile_base}
\end{figure}

\begin{figure}[t]
  \centering
  \includegraphics[width=\columnwidth]{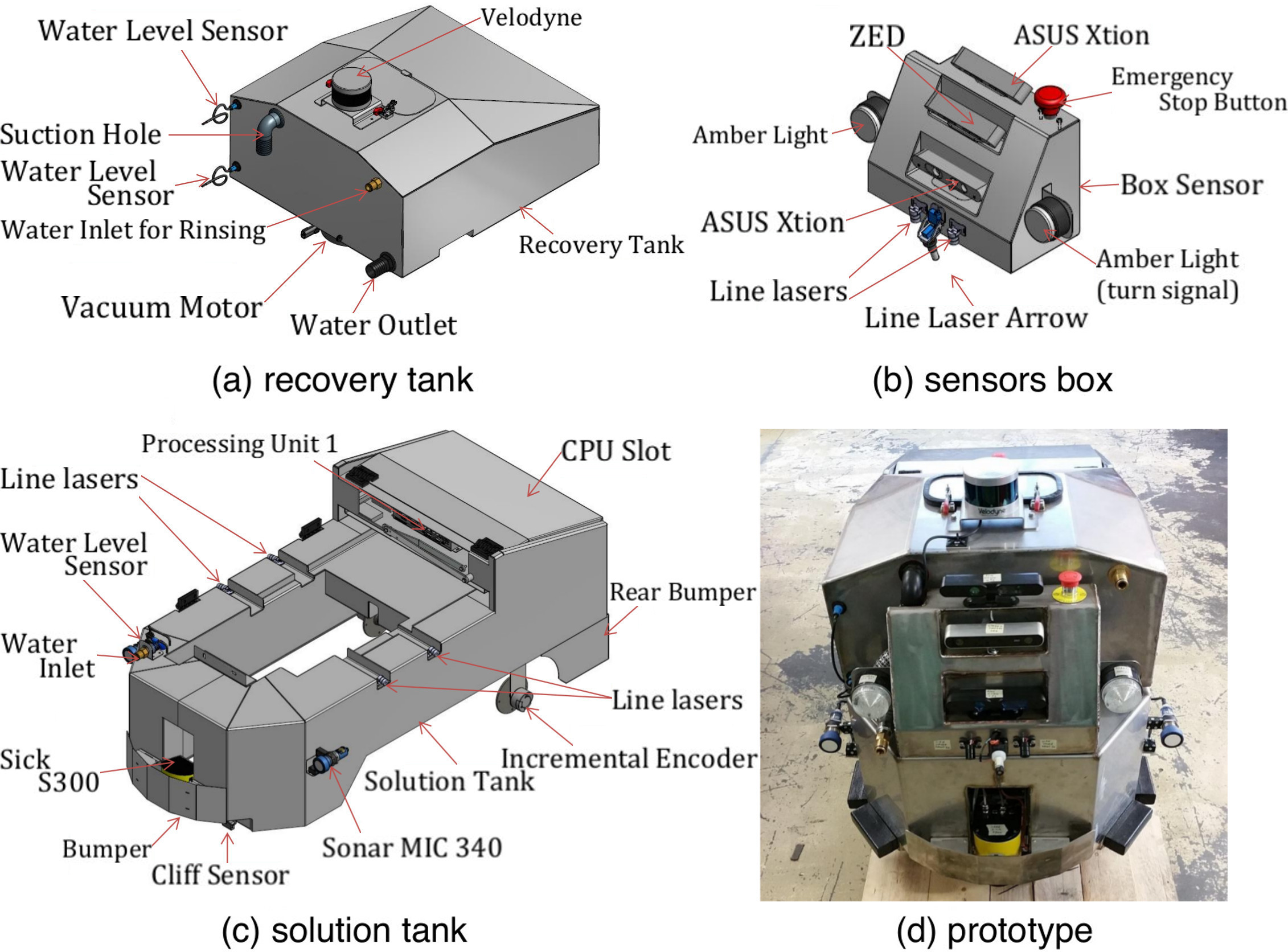}
  \caption{The sensor configuration of the FLOBOT prototype.}
  \label{fig:sensor_config}
\end{figure}

The mobility of FLOBOT is empowered by a typical three-wheeled base including two rear wheels powered by a single source and powered steering for the third (front) wheel, as shown in Fig.~\ref{fig:mobile_base}.
The sensor configuration is illustrated in Fig.~\ref{fig:sensor_config}.
Specifically, it includes:
\begin{itemize}
\item A 3D lidar (Velodyne VLP-16) is mounted at $0.8m$ from the floor, on the top of the robot.
  It captures a full 360$^\circ$ scene and generates point clouds of its surroundings.
  In order to adapt to its vertical field-of-view (30$^\circ$), we placed the sensor at the front of the robot and matched the streamlined design at the back to minimize occlusion.
  Although the effective detection distance of the lidar can reach approximately $100m$, as the distance increases, the point cloud will become increasingly sparse, which prevents human detection beyond $30m$.
  However, this distance has fully met the the safety requirements of FLOBOT.
\item Two RGB-D cameras (ASUS Xtion PRO LIVE), one facing forward and one facing the ground, are mounted at $0.55m$ and $0.72m$ from the floor, respectively, and used to detect human, dirt and objects.
\item A pointing downward stereo camera (ZED), mounted at $0.66m$ from the floor, is used as a complement to the floor-facing RGB-D camera.
  On surfaces with enough texture and in extremely bright situations its reliability was greater than the active RGB-D sensor, but its lack of precision meant that it was eventually omitted.
\item A 2D lidar (SICK S300) is mounted on the front of the robot, $15cm$ from the ground.
  It has a 270$^\circ$ horizontal field of view and a measurement range up to $30m$.
  As aforementioned, although its main use is in mapping and localization, its lower position is particularly suitable for human leg detection.
\item Two OEM incremental measuring wheel encoders are mounted on the outer cover (i.e. solution tank) of the robot and connected to the shafts of the rear wheels to obtain the robot's odometry.
\item An IMU (Inertial Measurement Unit, Xsens MTi-30) is installed in the front interior of the robot, horizontally placed above on z-axis of the front steering wheel.
  It provides the linear acceleration, angular velocity, and absolute orientation of the robot, and in combination with the odometry, the pose estimation of the robot itself can be greatly improved.
\end{itemize}
In addition to the above, other sensors include omni-directional trigger-bumpers, cliff sensors, and sonars.
Even though they are not directly connected to the perception software modules, there is an independent safety system triggering the emergency brake depending on the input, as well as the 2D lidar, which is the main purpose of using these sensors.

\begin{figure}[t]
  \centering
  \includegraphics[width=\columnwidth]{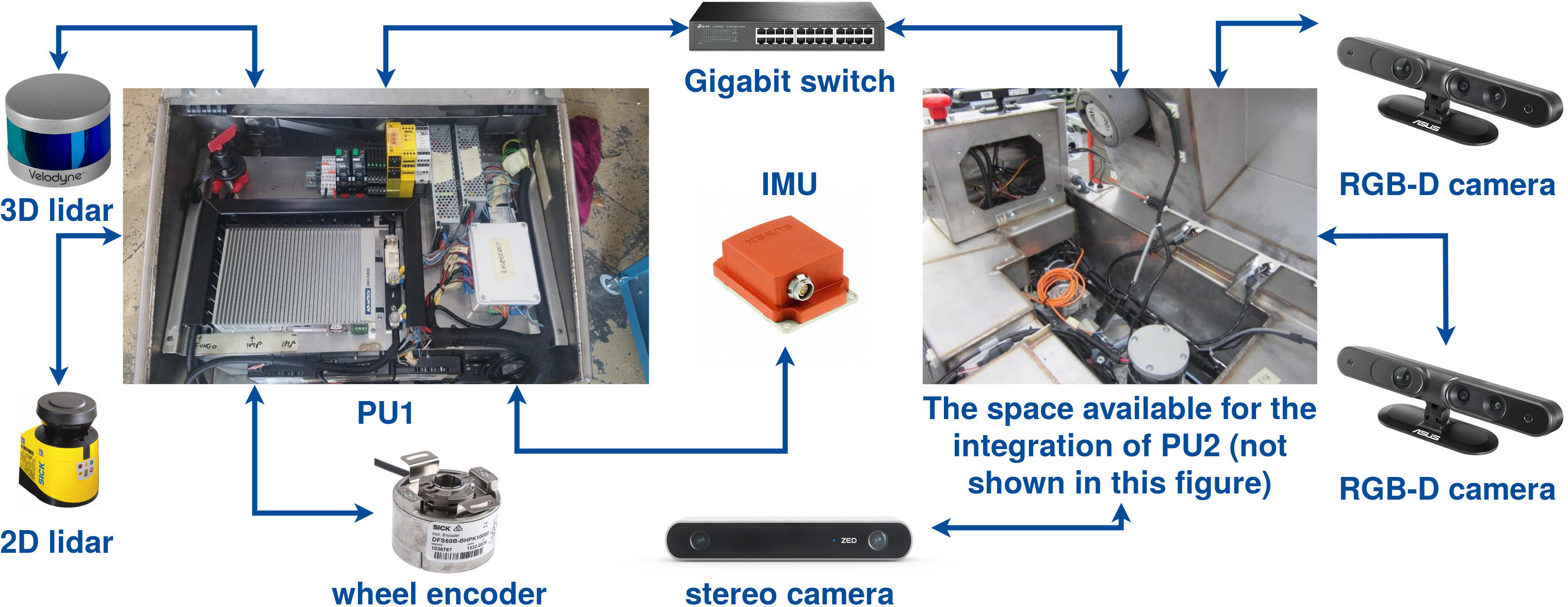}
  \caption{Connection diagram between sensors and computers.}
  \label{fig:hardware_connection}
\end{figure}

Processing Unit 1 (PU1), a passively cooled industry computer hosting the ROS core is used as master computer, which ensures operation of the most essential system modules such as sensor fusion, map-based navigation, 3D lidar-based human detection and tracking.
Processing Unit 2 (PU2), a consumer PC with a dedicated high-performance GPU serves as slave unit which is responsible to process computational intense and algorithmically complex jobs, especially for the visual computing such as dirt and floor object detection.
The communication between PU1 and PU2 is wired ensured by a Gigabit switch.
Regarding the network connectivity of the sensors (see Fig.~\ref{fig:hardware_connection}), the 3D and 2D lidars, the wheel encoder and IMU are wired connected to PU1, while the three cameras are connected to PU2.
In  addition, FLOBOT is equipped with a 104Ah Lithium battery that can provide about 2-3 hours of autonomy.

\subsection{Software architecture}

\begin{figure}[t]
  \centering
  \includegraphics[width=\columnwidth]{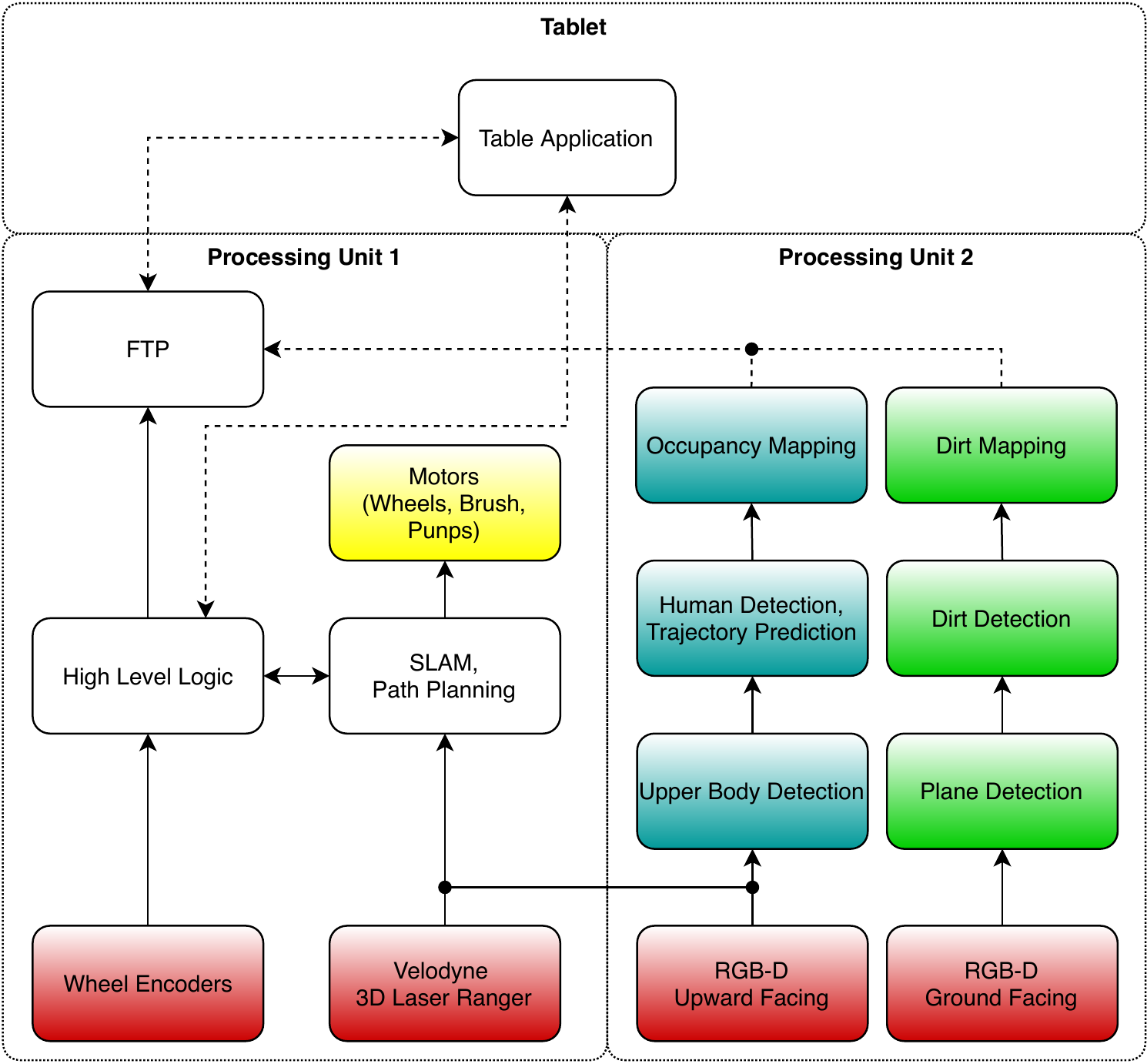}
  \caption{The FLOBOT software architecture. Solid lines are ROS based communication while dotted lines portray other methods.}
  \label{fig:flobot_software_overview}
\end{figure}

The FLOBOT software system is based entirely on ROS, a middleware designed with distributed computing in mind.
The software communication between PU1 and PU2 is therefore achieved through a ROS network consisting of a single ROS master and multiple ROS nodes.
The perception system consists of two parts: dynamic and static object detection.
The former mainly refers to humans, while the latter includes floor objects and dirt.
Details of the algorithms for navigation and ROS integration of perception modules are shown in Fig.~\ref{fig:flobot_software_overview}.

\section{Human detection and tracking}
\label{sec:dynamic_perception}

\begin{figure}[t]
  \centering
  \includegraphics[width=\columnwidth]{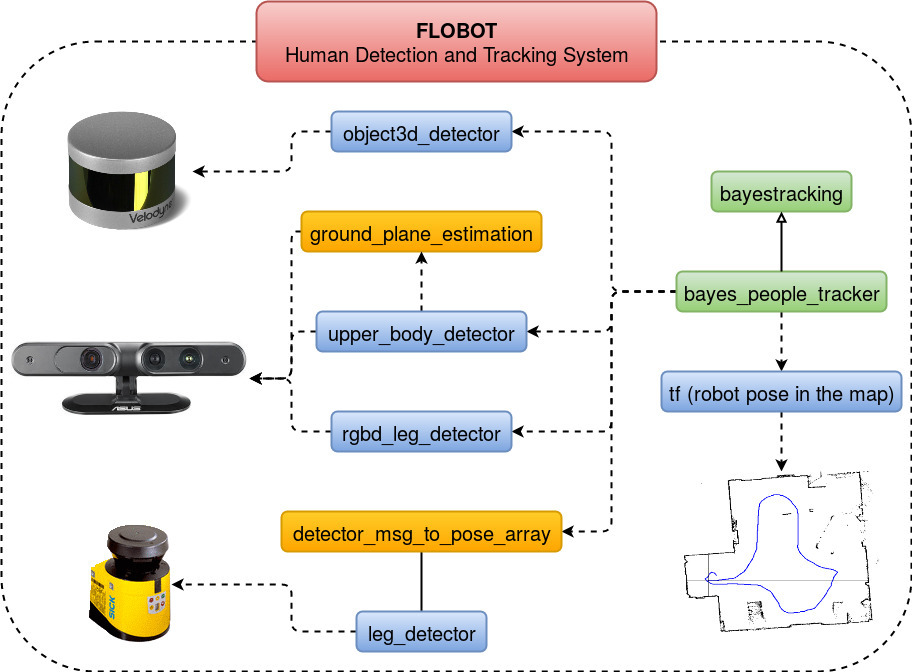}
  \caption{The UML diagram of the perception pipeline for human detection and tracking.}
  \label{fig:flobot_uol_uml}
\end{figure}

The human detection and tracking system simultaneously uses three different sensors to robustly track human movements in real time, and therefore increases the safety of the robot.
It fuses information about human location detected by the forward-facing RGB-D camera, the 2D and the 3D lidars, using Bayesian filtering~\citep{BayesianTracking}.
The system is robust enough thanks to the sensor configuration as well as the detection and tracking algorithms implemented.
In particular, the combined use of 2D and 3D lidars provides long-range and wide-angle detection, and additionally minimizes the perception occlusions, while the RGB-D camera is more reliable in the short range with accurate and robust algorithms.
The sensor location can be seen in Fig.~\ref{fig:sensor_config}, and a detailed view of the proposed system as a UML diagram is shown in Fig.~\ref{fig:flobot_uol_uml}.

\begin{figure}[t]
  \centering
  \includegraphics[width=\columnwidth]{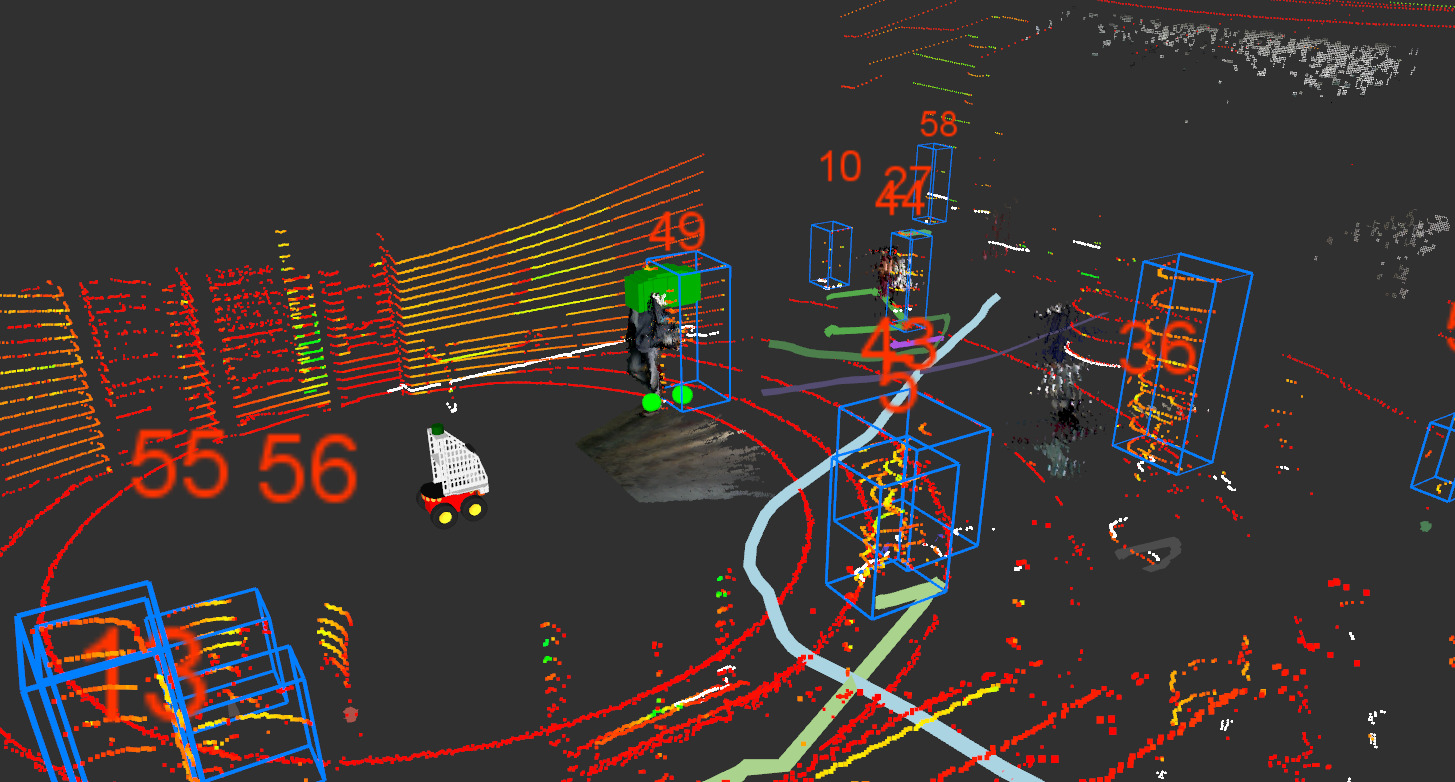}
  \caption{A screenshot of our multisensor-based detection and tracking system in action. The sparse colored dots represent the laser beams with reflected intensity from the 3D LiDAR. The white dots indicate the laser beams from the 2D LiDAR. The colored point clouds are RGB images projected on depth data of the RGB-D camera. The robot is at the center of the 3D LiDAR beam rings. The numbers are the tracking IDs and the colored lines represent the people trajectories generated by the tracker. For example, the person with tracking ID 49 has been detected by the RGB-D based \emph{upper\_body\_detector} (green cube), the 2D LiDAR based \emph{leg\_detector} (green circle), and the 3D LiDAR based \emph{object3d\_detector} (blue bounding box).}
  \label{fig:human_tracking}
\end{figure}

An initial version of the software was implemented on a MetraLabs Scitos G5 robot platform, in collaboration with researchers from another EU project STRANDS~\citep{STRANDS}.
The robot was equipped with sensors similar to the ones devised for the FLOBOT, i.e. a forward-facing RGB-D camera and a 2D lidar.
The former is used to detect the human upper body (i.e. \emph{upper\_body\_detector})~\citep{jafari14icra}, while the latter is used to detect human legs (i.e. \emph{leg\_detector})~\citep{arras07icra}.
In accordance with the FLOBOT requirements and specifications, in particular with the lower position of the RGB-D camera and the introduction of the 3D lidar, we have subsequently implemented two new human detection modules, i.e. an RGB-D camera-based leg detector (i.e. \emph{rgbd\_leg\_detector}) and a 3D lidar-based human detector (i.e. \emph{object3d\_detector}), and further improved the tracker (i.e. \emph{bayestracking} and \emph{bayes\_people\_tracker}) to adapt to long-distance large-volume people tracking.
Moreover, the two newly developed modules are based on PCL (Point Cloud Library)~\citep{PCL}, which is the state-of-the-art C\texttt{++} library for 3D point cloud processing.
For an intuitive understanding of the various detectors and their outputs, please refer to the example in Fig.~\ref{fig:human_tracking}.
The following paragraphs describe each module in detail.

\subsection{3D lidar-based human detector}

The 3D lidar-based human detector can be learned in either online~\citep{yz19auro,yz18iros,yz17iros} or offline manner.
For FLOBOT, the detector is based on a Support Vector Machine (SVM)~\citep{SVM}.
We evaluated the state-of-the-art SVM features for a 3D lidar-based human classifier~\citep{yz19auro} and selected several of them, combined with a new developed feature to improve classification performance according to the needs of FLOBOT.
The specific details are shown in Table~\ref{tab:svm_features}.
Seven features (a total of 71 dimensions) were used, of which $(f_1, \ldots, f_4)$ were introduced by~\citep{navarro-serment09fsr}, $f_5$ and $f_6$ were proposed by~\citep{kidono11iv}, while $f_7$ was presented by~\citep{yz19auro}.
Both online and offline modes train the classifier using LIBSVM~\citep{libsvm}.
For offline training, the ``L-CAS 3D Point Cloud Annotation Tool 2''\footnote{\url{https://github.com/yzrobot/cloud_annotation_tool/tree/devel}} can be used.
For the online case, please refer to our previous work~\citep{yz19auro,yz18iros,yz17iros} for more details.

\begin{table}
  \caption{Features used for 3D lidar-based SVM human classifier}
  \label{tab:svm_features}
  \begin{tabular}{lll}
    \hline\noalign{\smallskip}
    Feature & Description & Dim. \\
    \noalign{\smallskip}\hline\noalign{\smallskip}
    $f_1$ & Number of points included in the cluster & 1 \\
    $f_2$ & Minimum cluster distance from the sensor & 1 \\
    $f_3$ & 3D covariance matrix of the cluster & 6 \\
    $f_4$ & Normalized moment of inertia tensor & 6 \\
    $f_5$ & Slice feature for the cluster & 20 \\
    $f_6$ & Reflection intensity's distribution & 27 \\
    $f_7$ & Dis. from the centroid of each slice to the sensor & 10 \\
    \noalign{\smallskip}\hline
  \end{tabular}
\end{table}

Conventionally, the offline supervised learning techniques can guarantee the performance of the classifier.
However, labelling the training examples is tedious work which implies labor costs.
It is also to be expected that the classifier is required to be retrained with every change in sensor setup or when being introduced to a new environment, as expected for a product like FLOBOT.
We thus developed an online learning framework to not only adapt to different environments and allow the robot to update its human model on the fly, but also to compete with or exceed classifier performance of offline models.
Moreover, the online framework enables long-term robot autonomy, including the acquisition, maintenance and refinement of the human model and multiple human motion trajectories for collision avoidance and robot path optimization.

\subsection{RGB-D camera-based upper body detector}

A new RGB-D camera-based upper body detector was originally developed by the STRANDS~\citep{STRANDS} project and adapted for use in FLOBOT.
It uses a template and the depth information of the camera to identify upper bodies, i.e. shoulders and head~\citep{jafari14icra}.
To reduce the computational load, this detector employs ground plane estimation to determine a Region of Interest (RoI) most suitable to detect the upper bodies of a standing or walking person.
The actual depth image is then scaled to various sizes and the template is slid over the image trying to find matches.

\subsection{RGB-D camera-based leg detector}

The camera-based leg detector was developed to enhance the close-range human detection with the forward-facing RGB-D camera, mounted on the FLOBOT at $0.55m$ from the floor.
A cosine similarity approach is used, and the main steps of the detection process are illustrated in Algorithm~\ref{algo:rgbd_leg}.
Specifically, a registered RGB-D point cloud is first down-sampled to obtain fewer points to speed up subsequent processing.
The obtained point cloud is further processed by removing any planes contained, which further improves the efficiency of the pipeline, especially in indoor environments.
The remaining points are then segmented based on Euclidean distance and leg candidates are filtered according to a set of predefined rules.
Next, colour histograms of the candidates are calculated and any two of them are compared using the cosine similarity:
\begin{equation}
  {\text{similarity}}=\cos(\theta)={\mathbf{A} \cdot \mathbf{B} \over \| \mathbf{A} \|\| \mathbf{B} \|}={\frac {\sum \limits _{i=1}^{n}{A_{i}B_{i}}}{{\sqrt {\sum \limits _{i=1}^{n}{A_{i}^{2}}}}{\sqrt {\sum \limits _{i=1}^{n}{B_{i}^{2}}}}}}
\end{equation}
Finally, candidates with a strong similarity are considered a pair, while the closest pair within a certain distance are considered to be human legs.

\begin{algorithm}
  1. Downsampling incoming registered RGB-D point cloud using the PCL VoxelGrid filter;\\
  2. Removing all planes from the point cloud using the PCL plane segmentation;\\
  3. Segmenting the points at $0.55m$ from the ground using the PCL Euclidean Cluster Extraction;\\
  4. Filtering leg candidates according to the following rules:\\
  \qquad - Feet off the ground are no more than $0.2m$ high;\\
  \qquad - Legs are upright parallelepiped;\\
  \qquad - Legs are within a reasonable size (e.g. between $0.1m^3$ and $0.5m^3$);\\
  5. Calculating colour histogram (e.g. 64 bins) of the leg candidates;\\
  6. Calculating the cosine similarity between any two candidates;\\
  7. Labelling the closest pair of candidates as leg if their similarity is greater than the similarity threshold of $0.8$ and the Euclidean distance between them is less than $1.0m$.
  \caption{RGB-D camera-based leg detection using color histogram\label{algo:rgbd_leg}}
\end{algorithm}

Please note that in Algorithm~\ref{algo:rgbd_leg}, the parameter values are pre-defined empirical values set based on our experiments with the L-CAS dataset~\citep{yz17iros}.
The released source code allows users to enter different parameter values as needed to get the best performance according to their robot's operational environment.

\subsection{2D laser-based leg detector}

The 2D laser-based leg detector is part of the official ROS people stack\footnote{\url{https://github.com/wg-perception/people}} and was initially proposed by~\citep{arras07icra}.
It is very suitable for our use in FLOBOT because, similarly to the original paper, our robot has a 2D laser scanner located at $0.119m$ off the ground.
A set of 14 features has been defined for legs detection, including the number
of beams, circularity, radius, mean curvature, mean speed, and more.
These features are used for the supervised learning of a set of weak classifiers using recorded training data.
The AdaBoost algorithm is then employed to turn these weak classifiers into a strong classifier, detecting legs from laser range data.

\subsection{Bayesian tracker}

The Bayesian tracker was developed for robust multi-sensor people tracking, exploiting the rich information provided by the FLOBOT platform.
It extends and improves the solution proposed by~\citep{Bellotto2009}, which allows to combine multiple sensor data, independently from the particular detection type and frequency.
This tracker implementation is based on the UKF, which has been shown to achieve results comparable to a Sampling Importance Resampling (SIR) particle filter in several people tracking scenarios, but with the advantage of being computationally more efficient in terms of estimation time.
It is also possible to switch between UKF and SIR filters, or choose a standard Extended Kalman Filter (EKF), since they have all been implemented in the Bayesian tracking library.

In the current ROS implementation, different tracking configurations can be used by defining the noise parameters of a 2D Constant Velocity (CV) model to predict human motion.
Together with additional observation models this is used to compensate during temporary detection losses.
A gating procedure is applied using a validation region around each new predicted observation~\citep{nb18eor}, based on the chosen noise parameters, to reduce the risk of assigning false positives and wrong observations.
New validated detections are then associated to the correct target using a simple Nearest Neighbour (NN) data association algorithm or the more sophisticated and robust, but also computationally expensive, Nearest Neighbour Joint Probabilistic Data Association (NNJPDA).
Detections that have been consistently found within a specific time interval, but have not been associated to any existing target, are stored and eventually used to create new tracks.
For more details, please refer to~\citep{BayesianTracking,Bellotto2009}.

\section{Dirt and object detection}
\label{sec:static_perception}

\begin{figure}[t]
  \centering
  \includegraphics[width=0.52\columnwidth]{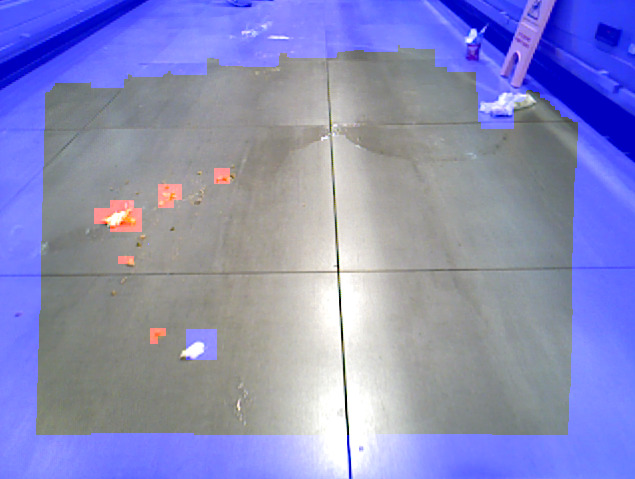}
  \includegraphics[width=0.468\columnwidth]{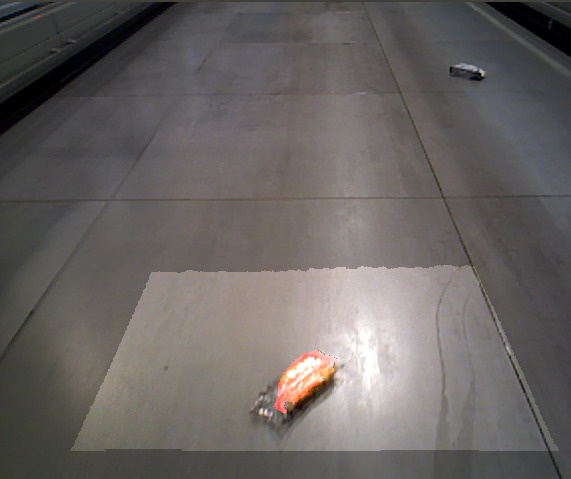}
  \caption{Detected rubbish (left) and object (right).}
  \label{fig:dirt_object_detection}
\end{figure}

The dirt and object detection module follows a simple pipeline (see Fig.~\ref{fig:flobot_software_overview}):
starting with a point cloud generated by the floor-facing RGB-D sensor we split up the data into floor and obstacles by a simple plane fitting approach;
the plane mask together with the RGB image provides the input for the dirt detection;
dirt detection then fits a model to the prevalent floor patterns and considers every outlier as dirt.
For an intuitive understanding of the dirt and object detection approaches, please refer to the example in Fig.~\ref{fig:dirt_object_detection}.
The following paragraphs describe each part in detail.

\subsection{Object detection}

Depending on the configuration of the cleaning equipment, it is beneficial to stop operation of the robot when objects appear in front of the robot.
If there is a rotating brush operating in front of the squeegee, most of the tiny objects would just be spun away, but in case there is a rubber lip, objects could get caught in it and interfere with the cleaning operation.
Since not all obstacles are caught by the relatively sparse lidar data, we see the use of the RGB-D sensor as an obvious solution in these cases. 

Conceptually, plane segmentation should be sufficient to differentiate floor from obstacles. Depending on the evenness of the floor, thresholds need to be adapted to make the plane model generous enough to handle deviations. 
In \citep{ss17oagm} we show that incorporating curvature into the floor model proves to be vital to overall performance.
We furthermore adopted a noise model for depth dependant thresholds. This enables us to put the thresholds extremely close to the sensor's noise level to detect objects which might only be protruding a few centimeters above the ground. Most objects higher than $2cm$ are detected at distances smaller than $1.3m$, which is sufficient for our application.

\subsection{Dirt detection}

Despite the FLOBOT project's premise to operate the robot in a multitude of environments, it was not expected to collect mission data until the final stages of the project.
Algorithms with long training phases and an appetite for a vast amount of domain-specific training data were therefore discarded in our considerations, and an unsupervised approach was selected.

Our algorithm (Algorithm~\ref{algo:dirt}) is based on the principle of novelty detection and driven by GMMs (Gaussian Mixture Models)~\citep{DudaHartStork01} trained on the gradient distribution of each input image.
The complexity of these GMMs is chosen such that they approximate a description of the currently visible floor but handle staining and spillage as outliers.

\begin{algorithm}
  1. Convert incoming RGB frame to Lab color space;\\
  2. Calculate the absolute value of gradient for channels;\\
  3. Split the image into blocks (of e.g. 16 by 16 pixel);\\
  4. Discard blocks which intersect with objects;\\
  5. Compute mean and standard deviation for gradient in each block:\\
  6. Train GMMs for each channel given mean and standard deviation of blocks as inputs;\\
  7. Predict Log-likelihood for each block based on GMM;\\
  8. Mix (sum) Log-likelihood of all channels;\\
  9. Labelling of pixel based on thresholding.
  \caption{GMM based dirt detection\label{algo:dirt}}
\end{algorithm}

There are some limitations attributed to this approach.
Specular highlights of various light sources will appear novel an therefore be misinterpreted as dirt.
Shadows of objects and people can be mistaken as dirt since they often appear to be isolated in a smaller portion of the image and therefore appear novel. 
Most of these effects are corrected in a dirt-mapping phase where observations are median filtered.
Specular highlights e.g. change its position during the robot's movement, while shadows of persons often shift quickly.
In those cases the filter will discard these measurements and only conserve static artifacts like dirt.

\section{Environment reasoning and learning}
\label{sec:reasoning_learning}

Since the potential mission areas of the robot are subject to changes, such as introduction and removal of dirt sources, we enable reasoning about the environment via a spatial-temporal map.
It takes four inputs including robot localization, trajectories of human beings, the cleanness measure and remaining dirt spots, and outputs the statistics of human trajectories and the dirt expectation distribution in the form of a heatmap.
This representation is intended to answer questions (what, when, where and how) such as:
What would be the best time for FLOBOT to clean, and where?
Should a given pollution be dry cleaned to avoid a slipping hazard, or is it necessary to apply a cleaning agent?

Answering these questions will need heuristics/algorithms that vary between different mission areas and customers.
While, for example, a warehouse with trained personnel might not care as much about slipping hazards, a wet cleaning mission during business times could be problematic in supermarkets.
Our solution enables future user studies/experience to formulate and implement the necessary behaviours.
In the following, we outline how human trajectories and the dirt heatmap representations are generated and discuss their expressiveness for future research and the robot's operation.

\begin{figure}[t]
  \centering
  \includegraphics[width=0.53\columnwidth]{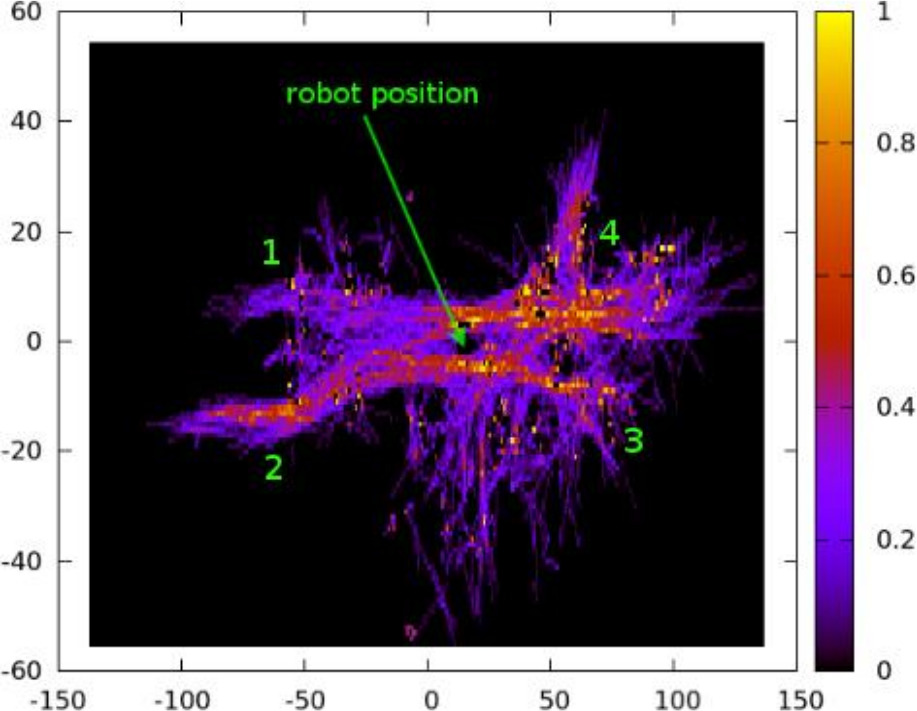}
  \includegraphics[width=0.45\columnwidth]{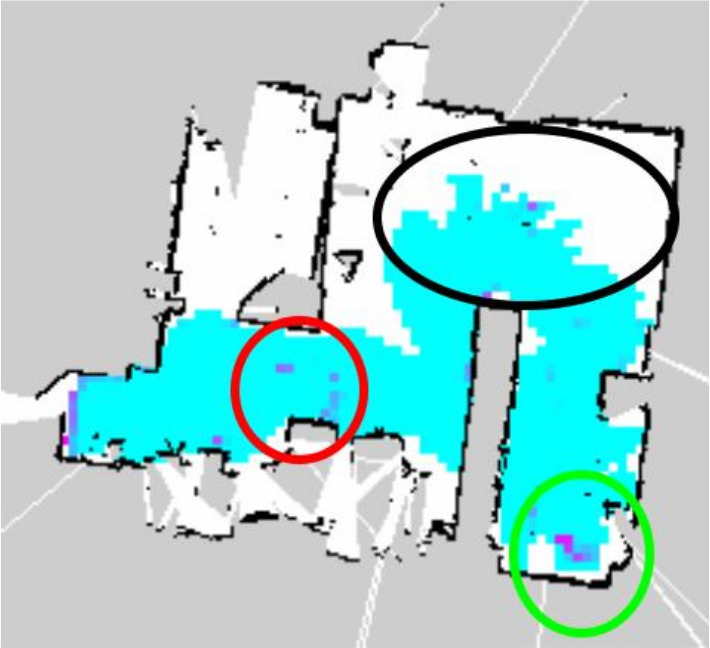}
  \caption{Left: human trajectories heatmap generated with the L-CAS 3D Point Cloud People Dataset. Warmer colors indicate higher frequencies of pedestrian occupancy. The map is normalized between 0 and 1. Right: dirt heatmap generated with the TUW Dataset. The circles indicate different dirt spots (false alarms included).}
  \label{fig:human_dirt_heatmaps}
\end{figure}

The human trajectories (i.e. sets of 2D coordinate) are generated with the system as presented in Sect.~\ref{sec:dynamic_perception}.
Based on the accumulated trajectories, a heatmap is generated to analyse the (context-related) characteristics of human activities.
For FLOBOT, it is actually an effective way to reflect the likelihood of human presence at a given site.
In particular, the trajectories are first discretized into a grid map with a cell size of $0.2m \times 0.2m$.
Then, the heatmap (see Fig.~\ref{fig:human_dirt_heatmaps}) is generated: the higher the number of trajectories passing by a cell, the brighter the colour, i.e. the higher the likelihood in the range $[0,1]$.

Based on Fig.~\ref{fig:human_dirt_heatmaps} (left), the following temporal-spatial analysis can be conducted.
The L-CAS data was recorded in a university atrium during lunch time (i.e. from 12AM to 1PM).
Zone 1 and 2 are both corridors with same width, but people were preferring to pass from zone 2, because there is a food shop over there.
Zone 3 and 4 are the liveliest places, as they are the entrance to the dining and food areas, respectively.
Consequently, an indicative decision that the FLOBOT can make would be ``it is better to clean zone 1 during lunch time''.
For path planning optimization, different maps for different times of the day could be further generated according to user needs.

Dirt detection, as described in Sect.~\ref{sec:static_perception}, is done by fitting a GMM to describe the pattern of the perceived floor.
Given a picture and a floor mask, the GMM is capable of delivering an estimate of where dirt might reside in this picture.
These estimates are passed through a temporal median filter to reduce false positives, and finally projected onto the map as an additional layer of information.
We specifically opted to store only the state of the floor as it is first perceived during a single mission to generate a status report of the area prior to cleaning.

\section{Evaluation}
\label{sec:evaluation}

\subsection{FLOBOT perception dataset}
\begin{figure*}[t]
  \centering
  \includegraphics[width=\textwidth]{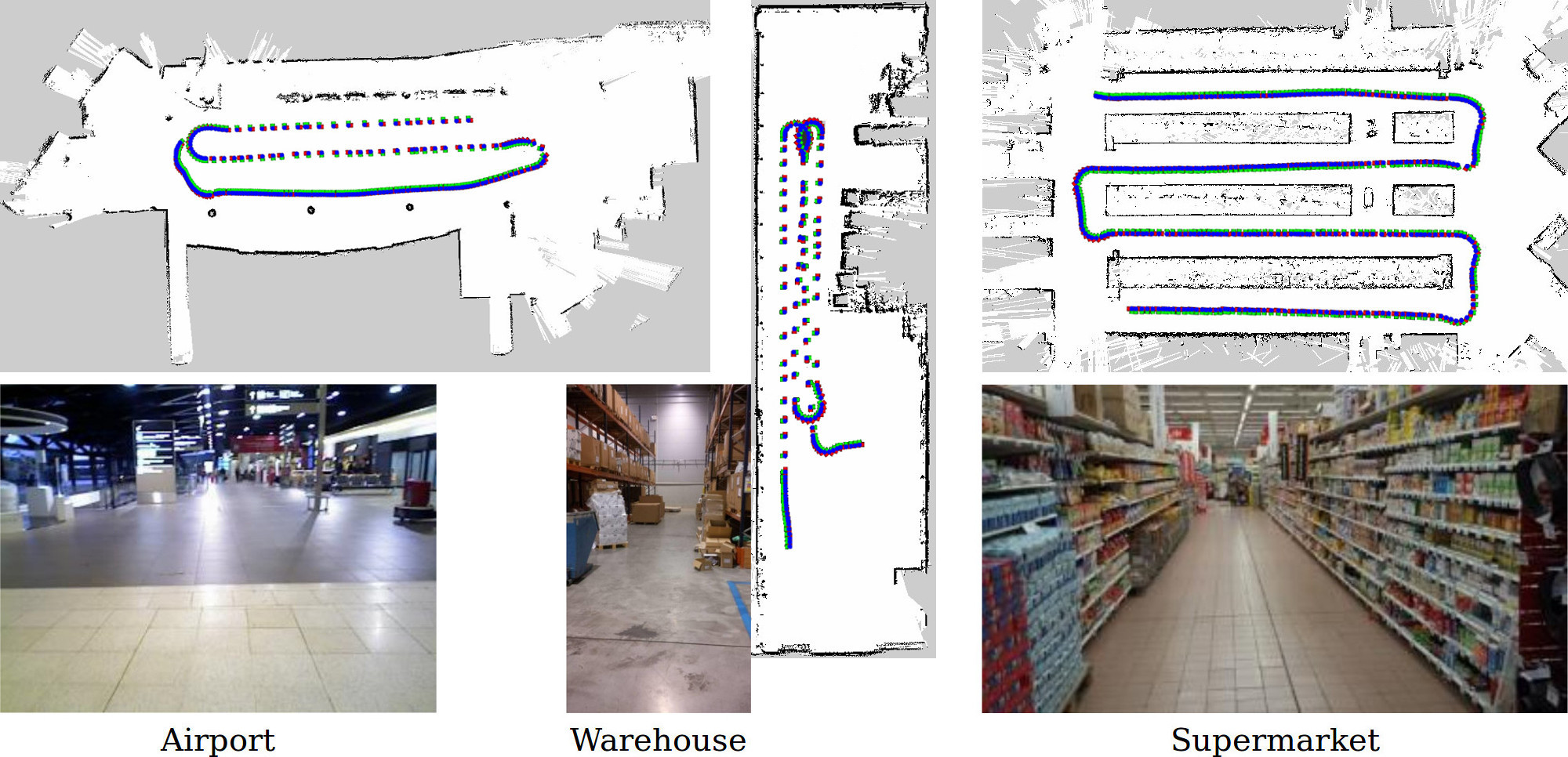}
  \caption{Three public places where the dataset recording was preformed. The upper part are occupancy grid maps generated from the Velodyne data, where the colored parts represent the footprints of FLOBOT.}
  \label{fig:dataset_overview}
\end{figure*}

The dataset recording was performed in three public places including an airport, warehouse and supermarket (see Fig.~\ref{fig:dataset_overview}), one in Italy and two in France.
Specifically, the Velodyne 3D lidar and the forward-facing depth camera\footnote{Please note that FLOBOT was not allowed to record any RGB data that can identify human identity information in the public places according to the EU General Data Protection Regulation (GDPR). Therefore, only depth information is allowed to be collected for the forward-facing Xtion PRO LIVE RGB-D camera.} data were collected for human detection and tracking purposes, while the floor-facing Xtion RGB-D camera data were collected for dirt and object detection purpose.
All sensory data, together with the robot pose in the world reference frame (i.e. ROS \emph{tf-tree} rising up to ``world''), were synchronized at the software level (i.e. time stamped by ROS) and recorded into several ROS \emph{rosbags}, according to their purpose and recording time.
The dataset is publicly available at \url{http://lcas.github.io/FLOBOT/}, and the relevant data statistics are shown in Table~\ref{tab:dataset_statistics}.

\begin{table*}
  \caption{Data statistics of the FLOBOT perception dataset.}
  \label{tab:dataset_statistics}
  \begin{tabular}{llll|l}
    \hline\noalign{\smallskip}
    Date & Time (GMT+2) & Place (Europe) & \multicolumn{2}{l}{Number of frames}\\
    \noalign{\smallskip}\hline\noalign{\smallskip}
    2018-04-19 & 11:41-11:49 (8:24s) & Carugate (supermarket) & 5,042 Velodyne & \multirow{2}{*}{2,174 Xtion (floor)}\\
    2018-05-31 & 16:35-16:39 (3:44s) & Carugate (supermarket) & 2,248 Velodyne / 6,729 Xtion (forward) &\\
    \hline
    2018-06-12 & 17:10-17:13 (3:27s) & Lyon (warehouse) & 2,073 Velodyne / 6,204 Xtion (forward) & \multirow{4}{*}{14,580 Xtion (floor)}\\
    2018-06-13 & 16:11-16:17 (5:05s) & Lyon (airport) & 3,059 Velodyne / 9,158 Xtion (forward) &\\
    2018-06-13 & 16:20-16:23 (2:26s) & Lyon (airport) & 1,460 Velodyne / 4,366 Xtion (forward) &\\
    2018-06-13 & 16:37-16:42 (4:28s) & Lyon (airport) & 2,688 Velodyne / 8,047 Xtion (forward) &\\
    \noalign{\smallskip}\hline
  \end{tabular}
\end{table*}

\subsubsection{Human detection and tracking}

\begin{figure}[t]
  \centering
  \includegraphics[width=\columnwidth]{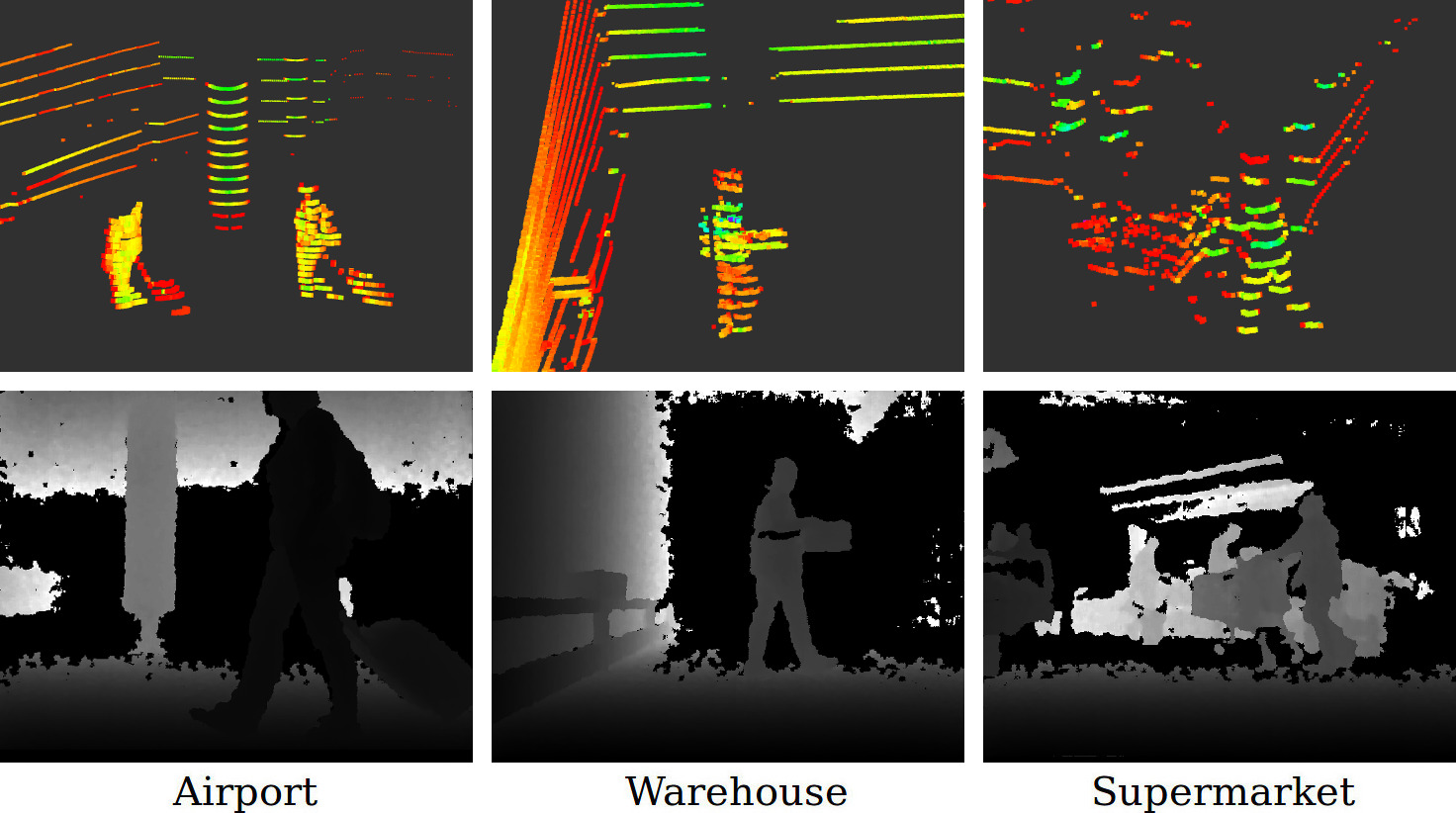}
  \caption{Scene-related human presentation in the FLOBOT dataset. The top half is 3D lidar data, while the lower part is the depth camera data.}
  \label{fig:human_dataset}
\end{figure}

Our dataset contains challenges in human detection and tracking, in particular caused by the scene-related human representation with the 3D lidar point clouds.
As shown in Fig.~\ref{fig:human_dataset}, passengers at the airport are typically carrying luggage, warehouse staff usually carry goods, and shoppers in the supermarket are normally pushing trolleys.
Besides these scene-related activities, staff from the research team also acted as pedestrians moving around the robot, for the purpose of module evaluation.

\subsubsection{Dirt detection}

\begin{figure}[t]
  \centering
  \includegraphics[width=0.32\columnwidth]{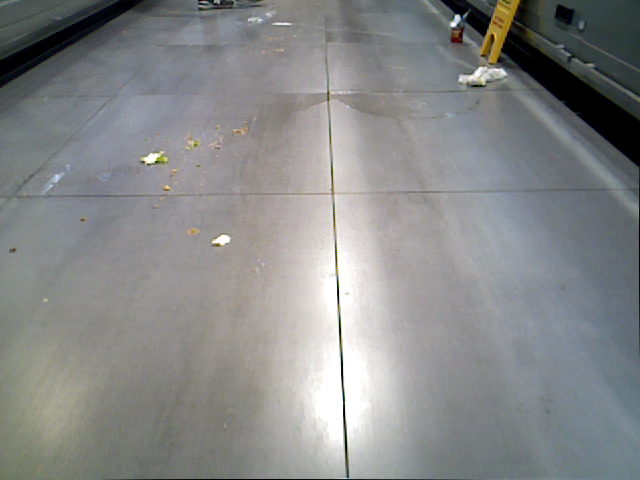}
  \includegraphics[width=0.32\columnwidth]{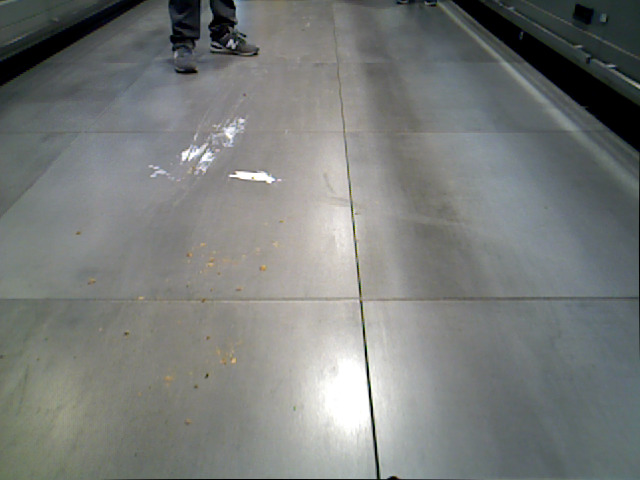}
  \includegraphics[width=0.32\columnwidth]{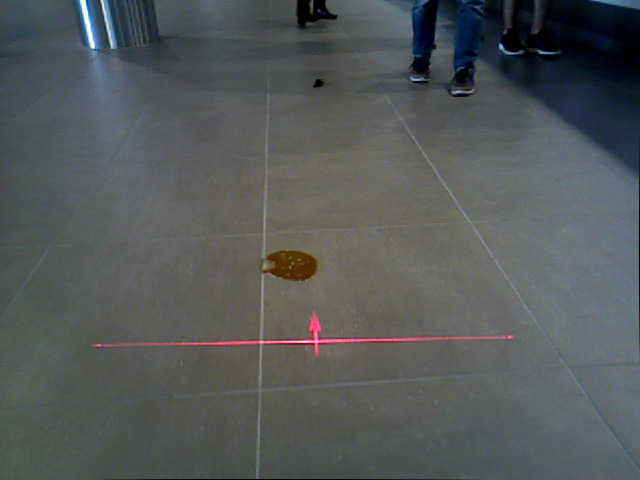}\\
  \vspace{0.1cm}
  \includegraphics[width=0.32\columnwidth]{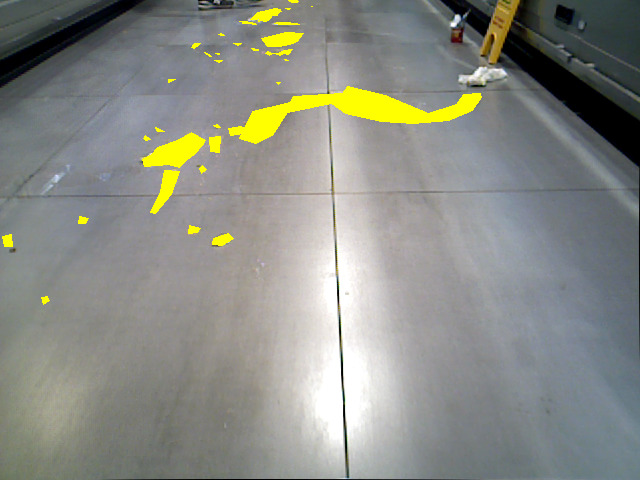}
  \includegraphics[width=0.32\columnwidth]{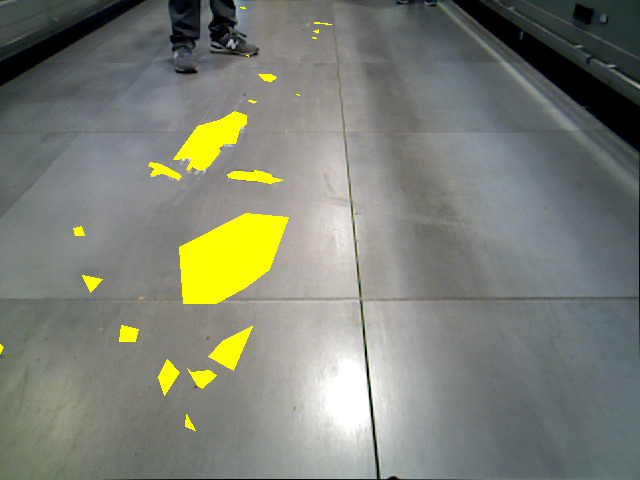}
  \includegraphics[width=0.32\columnwidth]{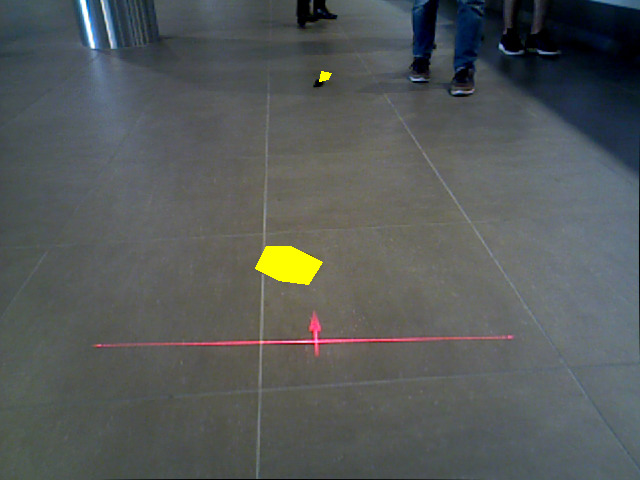}
  \caption{Some frames taken from our dirt dataset. The top row shows the raw frames, the bottom row shows the masked dirt in yellow. The two columns on the left depict frames captured in a supermarket, while the rightmost column is captured during a mission in a open space in an airport. During this mission the proactive safety module was active. Since its red laser markings would interfere with dirt detection, it is masked out in that processing step. The dataset provides masks for dirt, floor and said laser markings.}
  \label{fig:dirt_dataset}
\end{figure}

For evaluation purposes, we constructed pollution scenarios with materials found on site.
For example, in the supermarket, we contaminated the mission area with expired products such as milk, juice and cookies (which are well spread over the place), while a can of coke was used as a source of pollution in the airport.
Both scenarios are featured in tracks with pollution being annotated as polygons.
Annotations were performed with our Python-based annotation tool\footnote{\url{https://github.com/SimonTheVillain/flobotAnnotator}}. Given \emph{rosbags} as input, it enables us to label planar regions with polygons.
To reduce labour, the tool can propagate the labels (i.e. polygons) between frames according to the localization system running on the robot, via the \emph{tf-tree} between frames.
To overcome any inaccuracies in the trajectory, miscalibration and other issues, our tool also provides the option to move the position of a mask between keyframes.
Moreover, to keep the dataset and its usage as simple as possible we provide the captured frames in a PNG format as well as the masks for dirt, floor\footnote{Based on our plane estimation.} and when applicable, the mask for the projected laser markings.
Some frames taken from our dirt dataset can be seen in Fig.~\ref{fig:dirt_dataset}.
Ultimately this dataset is, by its size and diversity, not sufficient to train CNNs, but rather intended to serve as a validation dataset for the task at hand. 

\subsection{Results}

Together with the new dataset, we also open-source the aforementioned human detection and tracking\footnote{\url{https://github.com/LCAS/FLOBOT}} and dirt and object detection\footnote{\url{https://owncloud.tuwien.ac.at/index.php/s/h8ZDeypUJoRFmb4}} systems.
Some key modules were tested on the dataset to serve as baselines for further research.
We show experiments outside the laboratory, i.e. on a real prototype in real environments such as airport, warehouse and supermarket.
Below we give the relevant details.

\subsubsection{Human detection and tracking}

We provide a pre-trained SVM model for 3D lidar-based human detection and tracking to the community, which is publicly available together with the released system.
It is a binary SVM-based classifier (i.e. human or non-human) trained with 968 positive (i.e. human) examples and 968 negative (i.e. background) examples from the L-CAS dataset\footnote{File name: LCAS\_20160523\_1239\_1256\_labels.zip} \citep{yz17iros}.
The positives are manually annotated while the negatives are randomly selected from point clusters that are not human.
Technically, the LIBSVM~\citep{libsvm} is used for training with the aforementioned seven features (c.f. Table~\ref{tab:svm_features}), while all the feature values are scaled within the interval $[-1, 1]$.
The SVM model uses a Gaussian Radial Basis Function (RBF) kernel~\citep{SVM} and outputs probabilities associated to the labels.
In order to find the optimal training (best fitting) parameters, a five-fold cross validation is used for parameter tuning, especially for the cost of constraints violation and $\gamma$ in kernel function.

\begin{figure}[t]
  \centering
  \includegraphics[width=\columnwidth]{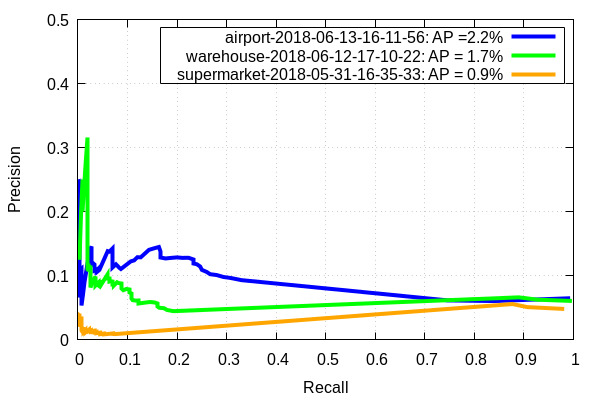}
  \includegraphics[width=\columnwidth]{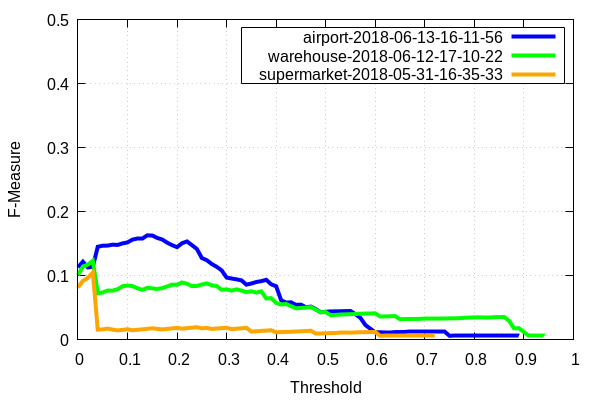}
  \caption{Evaluation of the generalization ability of the offline-trained human classifier. Test sets are built according to the traditional training-test $7:3$ ratio, i.e. 415 randomly selected examples for each scene of the FLOBOT dataset. The classification performance is evaluated using Precision, Recall, Average Precision (AP) and F-measure.}
  \label{fig:svm}
\end{figure}

The evaluation of our clustering algorithm, as well as human classifiers (trained either in offline or in online manner) for the same environment, can be found in our previous work~\citep{yz19auro,yz18iros,yz17iros}, while that of our tracking system can be found in~\citep{linder16icra,BayesianTracking}.
In this paper, we are more interested in the generalization ability of our system, as data for different environments are available.
Experimental results (see Fig.~\ref{fig:svm}) show that the generalization ability of the offline-trained classifier is extremely limited, i.e., training with data collected in a university atrium (the L-CAS dataset), while evaluating with data collected in an airport, a warehouse and a supermarket (the FLOBOT dataset).
This is mainly because not only the features of negative examples (i.e. background) are not similar, but also the differences of positive examples (i.e. human).
A typical example is that the dress code of a worker in the warehouse results in a significant difference of the point cloud intensity (the most representative feature for human classification) from the normal clothes.

However, our human-like volumetric model proposed in \citep{yz17iros} exhibits interesting results, as shown in Table~\ref{tab:detection_evaluation}.
The model serving as prepossessing of the human classification, is formulated as follows:
\begin{equation}
  \label{eq:volume_filter}
  \begin{aligned}
    HumanCandidate~=~\{C_i~| & 0.2 \leq w_i \leq 1.0,\\
    & 0.2 \leq d_i \leq 1.0,\\
    & 0.5 \leq h_i \leq 2.0\}
  \end{aligned}
\end{equation}
where $w_i$, $d_i$ and $h_i$ represent, respectively, the width, depth and height (in meters) of the cluster volume.
Together with our clustering algorithm, which divides the 3D space into nested circular regions centred at the sensor (like wave fronts propagating from a point source), additionally separating different objects and leading to the very promising results in Table~\ref{tab:detection_evaluation}.

\begin{table}
  \caption{Detection results on the FLOBOT dataset (airport, warehouse and supermarket) and L-CAS dataset (university).}
  \label{tab:detection_evaluation}
  \begin{tabular}{lllll}
    \hline\noalign{\smallskip}
    & Accuracy & Precision & Recall & F1-measure\\
    \noalign{\smallskip}\hline\noalign{\smallskip}
    Airport & 0.89 & 0.38 & 0.84 & 0.52\\
    Warehouse & 0.94 & 0.48 & 0.92 & 0.63\\
    Supermarket & 0.90 & 0.31 & 0.85 & 0.45\\
    University & 0.88 & 0.33 & 0.88 & 0.48\\
    \noalign{\smallskip}\hline
  \end{tabular}
\end{table}

\begin{figure}[t]
  \centering
  \includegraphics[width=\columnwidth]{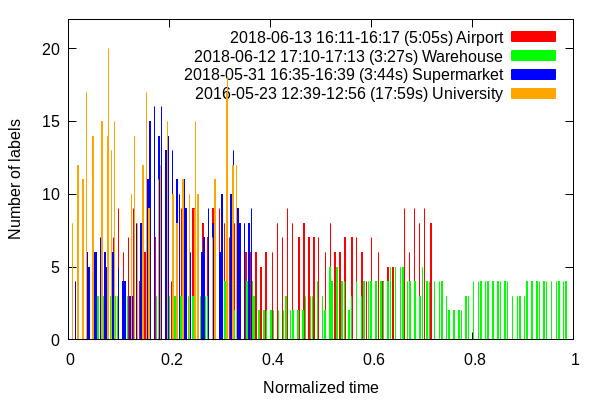}
  \caption{Human label distribution statistics of our test set. Best viewed in color. The university and supermarket data contain the most human labels per frame due to its large scene and its nature as a place of human gathering. The warehouse data has the fewest human labels per frame, due to its small scene and being open only to staff. The airport data contains a moderate number of human labels per frame because we selected a non-busy area to avoid passenger inconvenience.}
  \label{fig:labels}
\end{figure}

We randomly extract some frames from each scene data and fully annotate them to obtain 415 positive sample labels\footnote{The labels are available on the dataset website, annotated by using our open source annotation tool \url{https://github.com/yzrobot/cloud_annotation_tool/tree/devel}.} for each scene (label distribution as shown in Fig.~\ref{fig:labels}) and use them as the test set.
For the evaluation, we calculate the Intersection over Union (IoU) of two 3D bounding boxes, i.e. between the manually annotated ground truth and the human candidates, and the IoU threshold is set to $0.5$.
It can be seen from Table~\ref{tab:detection_evaluation} that,
$1)$ overall, the accuracy of the detector is high because the proportion of negative samples in all scenes is large;
$2)$ the precision is low as many negative examples have a human-like volume and are incorrectly detected as false positives;
$3)$ high recall with low precision actually shows an important trade-off we made for FLOBOT, i.e. since the robot is for professional users, it is expected to not miss any humans but can have false positives within a reasonable range;
$4)$ the best results are shown in the warehouse scenario, while the worst are in the supermarket. The former has a relatively simple environment and a small number (five) of people, while the latter is quite complicated and has a large number of shoppers.
This also shows that the performance of the detector is limited by the complexity of the environment.

\subsubsection{Dirt detection}

The environments the robot was operated in offered different types of floor and lighting conditions.
Some of these are challenging due to broken tiles, worn through coating, stains of paint, markings, drain gates and similar.
Even with a perfect novelty detection, these situations would not be solved, which reinforces our strong belief that learning based methods are the key to reliably operate in such applications.

\begin{figure}[t]
  \centering
  \includegraphics[width=0.32\columnwidth]{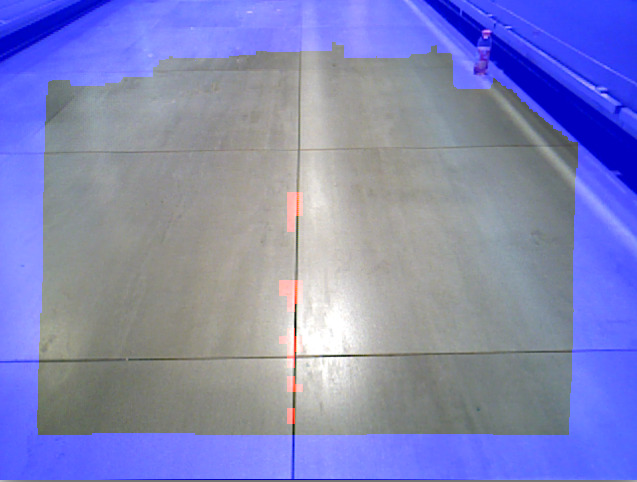}
  \includegraphics[width=0.325\columnwidth]{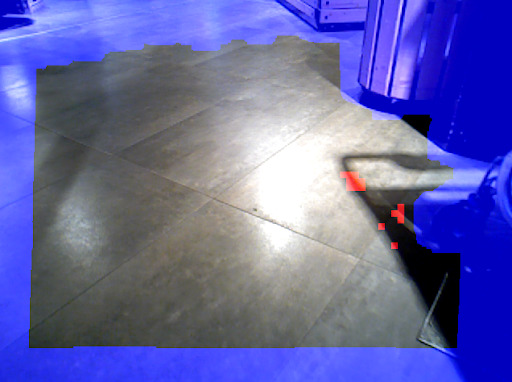}
  \includegraphics[width=0.323\columnwidth]{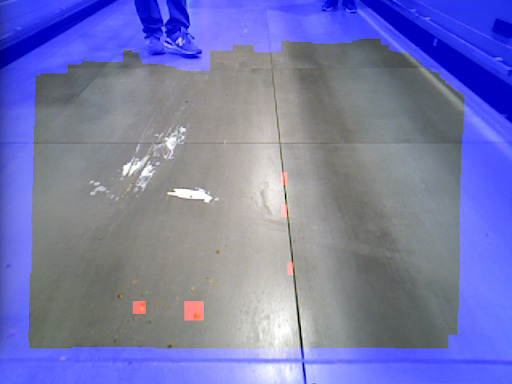}
  \caption{Scenarios challenging the novelty detection. Dirt is marked red whereas blue are pixels that are not considered. The floor tiles are just prominent enough not to be fitted into the GMM and considered as dirt (left). The same applies to shadows (middle). The other extreme occurs when the GMM generalizes too much and thus also incorporates dirt into its floor model (right).}
  \label{fig:gmm_fails}
\end{figure}

The ACIN dataset used in~\citep{ag17taros} only poorly reflects the challenges found in supermarkets.
Even though the proposed algorithm proves to be powerful on the said dataset, applying it on the data collected on site immediately exposes its deficiencies (see Fig.~\ref{fig:gmm_fails}).
Prominent gaps between tiles, specular highlights, sharp shadows and dirt with similar color as the floor all pose a challenge to novelty detection.
This is something that needs to be addressed in a modern dataset.
In our new airport and supermarket data, we created scenarios with spillages of goods available on site.
Arguably these are still imposed situations but the circumstances and used products make it more challenging and life-like than the reference datasets.

\begin{figure}[t]
  \centering
  \includegraphics[width=\columnwidth]{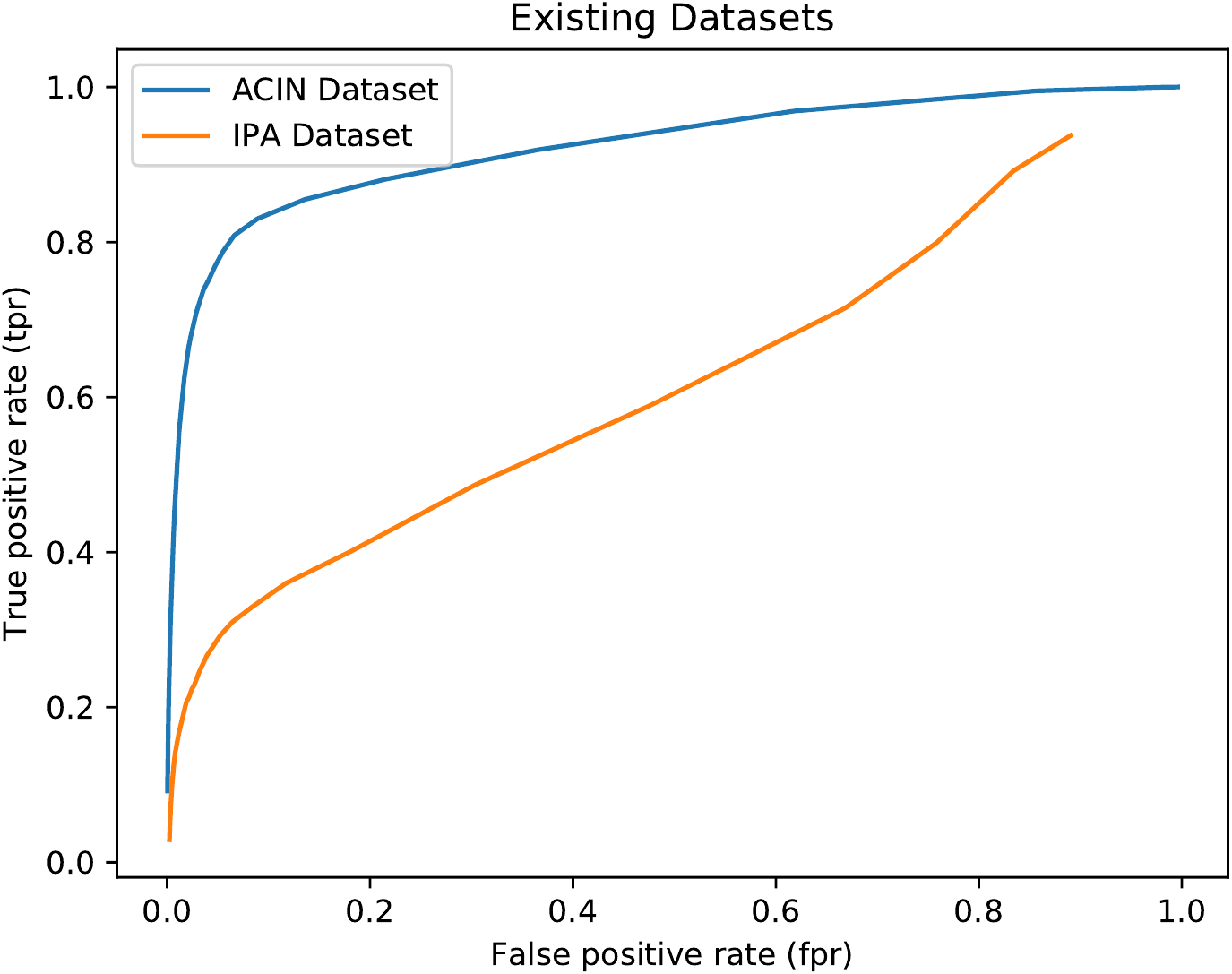}
  \includegraphics[width=\columnwidth]{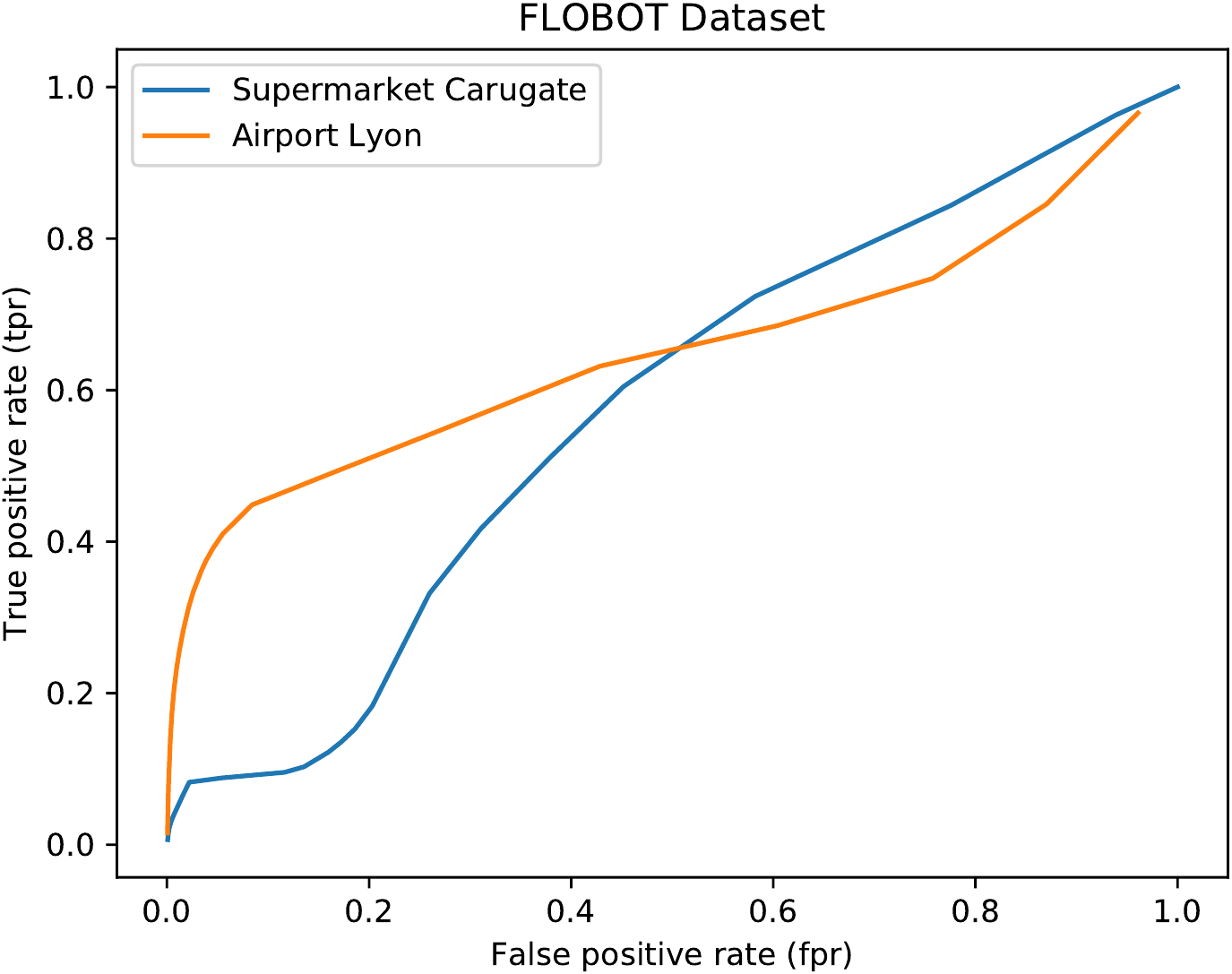}
  \caption{The algorithm presented in~\citep{ag17taros} performs favorably on old, lab-grown datasets (up). The data collected on site in a supermarket as well as an airport paints a different picture (down) as the same algorithm disappoints.}
  \label{fig:roc_dirt_datasets}
\end{figure}

To provide a baseline of the dirt-detection itself we decided to directly evaluate the algorithm without the median filter our pipeline uses downstream.
Fig.~\ref{fig:roc_dirt_datasets} gives a genuine indication of the core algorithm's capabilities.
While the algorithm performs reasonably on the ACIN dataset, it fails on the other datasets.
Taking the IPA dataset~\citep{rb13icra} as a comparison, we see data created in similar environments but with different post processing.
We argue that the annotations are narrower to the actual dirt, which makes it hard for our algorithm to perform favorably when calculating IoU.
For the datasets captured by FLOBOT, the annotations are even tighter by utilizing filled polygons instead of rectangles.

\section{Conclusions}
\label{sec:conclusions}

In this paper, we presented a robot perception system for an autonomous floor scrubber, including in particular the human detection and tracking module, the dirt and object detection module, and the combined use of the two within the environment learning and reasoning module.
The human detection and tracking module has been developed to enable safer robot navigation among humans by robustly and accurately tracking multiple people in real time.
The algorithm as stated in~\citep{ag17taros} for dirt spot detection is state of the art.
But even then, its results are solely to be interpreted by the operator on the tablet.
We have shown that areas of pollution are clearly visible in this representation with very few false positives surpassing filtering and ending up in the map.
Our claim is that, given this information, cleaning missions can be planned more efficiently.
We hope that, with increasing reliability of dirt detection algorithms, it will be possible for cleaning robots to make decisions more in line with the expectations of human operators.

The new dataset we collected is a valid addition to the existing ones~\citep{yz18iros,yz17iros,ag17taros,rb13icra}.
It provides out-of-lab data including airport, warehouse and supermarket environments, in which people usually have different clothes, belongings, and gaits in different public places, providing significant challenges for human detection and tracking.
It adds two new floor types and offers a variety of dirt and spillages, while offering increased difficulty due to specular reflections, shadowing and more prominent tile-gaps.
Deep-learning based methods hold great potential for these tasks but will need vastly more training data than collected here.
Extensive data collection together with artificial renderings will be needed to bridge this gap.

The results are first steps towards future autonomous service robots that work more independently and continue learning.
It could be envisioned that the robot keeps collecting samples where decisions are unclear, to let a user make a few clicks to improve the adaptation to a specific environment.
This would allow the cleaning machine to optimise its operation over time in a given environment, improving productivity and upskilling of cleaning professionals.
We also anticipate the adoption of similar methods in many other applications of service robots in human environments.

\begin{acknowledgements}
The authors would like to thank Fimap SpA for providing Fig.~\ref{fig:mobile_base} and \ref{fig:sensor_config}.
\end{acknowledgements}

\section*{Conflict of interest}
The authors declare that they have no conflict of interest.

\bibliographystyle{spbasic}
\bibliography{jfrExampleRefs}

\end{document}